\documentclass[letterpaper, 10 pt, conference]{ieeeconf}  

\IEEEoverridecommandlockouts                              

\usepackage{amsfonts}
\usepackage{amsmath}
\usepackage{graphicx}

\usepackage{mathtools}
\DeclarePairedDelimiterX{\infdivx}[2]{(}{)}{%
	#1\;\delimsize\|\;#2%
}
\newcommand{\infdiv}{\infdivx}

\usepackage{xargs}
\usepackage[pdftex,dvipsnames]{xcolor}  

\usepackage[
backend=bibtex,
style=numeric,
sorting=none
]{biblatex}
\usepackage{algorithm}
\usepackage{algpseudocode}

\title{\LARGE \bf
Diversity for Contingency: Learning Diverse Behaviors \\ for Efficient Adaptation and Transfer
}

\author{Finn Rietz$^{1}$ and Johannes A. Stork$^{1}$
\thanks{*This work was partially supported by the Wallenberg AI, Autonomous Systems and Software Program (WASP) funded by the Knut and Alice Wallenberg Foundation.}
\thanks{$^{1}$Adaptive and Interpretable Learning Systems Lab, Center for Applied Autonomous Sensor Systems, \"Orebro University, Sweden, {Correspondence: \tt\small finn.rietz@oru.se}}%
}

\begin{document}

\maketitle
\thispagestyle{empty}
\pagestyle{empty}

\begin{abstract}
Discovering all useful solutions for a given task is crucial for transferable RL agents, to account for changes in the task or transition dynamics.
This is not considered by classical RL algorithms that are only concerned with finding the optimal policy, given the current task and dynamics. 
We propose a simple method for discovering all possible solutions of a given task, to obtain an agent that performs well in the transfer setting and adapts quickly to changes in the task or transition dynamics. 
Our method iteratively learns a set of policies, while each subsequent policy is constrained to yield a solution that is unlikely under all previous policies.
Unlike prior methods, our approach does not require learning additional models for novelty detection and avoids balancing task and novelty reward signals, by directly incorporating the constraint into the action selection and optimization steps. 
\end{abstract}

\section{INTRODUCTION}
\begin{figure}[t]
	\includegraphics[width=0.5\textwidth]{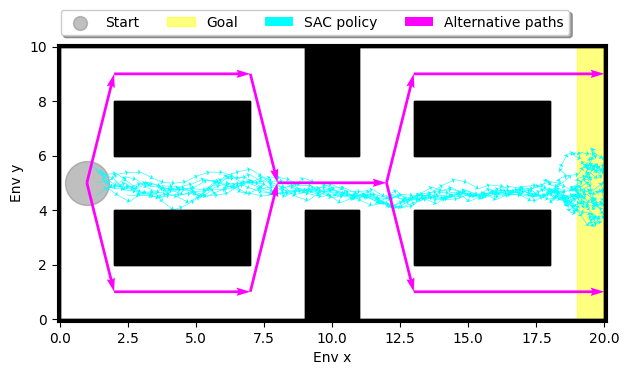}
	\centering
	\caption{Rollouts generated by a soft actor-critic agent. The behavior emits local stochasticity but does not learn the alternative paths. Discovering all possible solutions is crucial for transfer and adaptation.}
	\label{fig:rollouts}
\end{figure}

The standard reinforcement learning (RL) approach~\cite{sutton2018reinforcement} learns deterministic policies~\cite{schulman2017proximal, schulman2015trust, lillicrap2015continuous} for each task from scratch, despite the notorious sample inefficiency of deep RL algorithms. 
Instead, it would be preferable to learn transferable and reusable policies and to adapt them to different downstream tasks, with a fraction of data and compute needed compared to learning from scratch. 
A promising approach for learning transferable RL agents is multi-objective RL (MORL), where vectorized value functions can be shared for many tasks~\cite{haarnoja2018composable, hunt2019composing, barreto2017successor}.
A key requirement for learning such transferable agents is to allow stochasticity and diversity in the learned behavior~\cite{zhang2019novelPolicies, hausman2018learning}, as opposed to learning one overly specific, deterministic policy. 
While MaxEnt RL~\cite{ziebart2008maximum, haarnoja2017reinforcement, haarnoja2018composable} regularizes policies in an attempt to prevent them from becoming overly specific, entropy-regularized (MO) RL is not sufficient for inducing agents that learn all behaviors that solve the given tasks, as can be seen in Fig.~\ref{fig:rollouts}. 
To adapt transferred agents efficiently it is important to discover \textit{all} useful behaviors, to account for possible contingencies, e.g. parts of the middle pathway becoming blocked.

In this paper, we first review MaxEnt RL and methods that learn diverse behaviors, either unsupervised or for a given task. 
In Sec.~\ref{sec:method}, we then propose a novel method for learning policies that discover different solutions for the given task, accounting for possible contingencies in transfer settings.

\section{Related work}\label{sec:rel_work}
\subsection{Entropy regularization}
A common approach to learning stochastic policies with wide and smooth maxima is MaxEnt RL~\cite{haarnoja2017reinforcement, ziebart2008maximum}.
Maximum entropy (MaxEnt) RL augments the RL objective by adding a term proportional to the policy's entropy to the reward
\begin{equation}
		J(\pi) = \sum_{t=1}^\infty \mathbb{E}_{(\mathbf{s}_t, \mathbf{a}_t)} \bigg[ \gamma^{t-1} r(\mathbf{s}_t, \mathbf{a}_t) + \alpha \mathcal{H}\big(\pi(\cdot \mid \mathbf{s}_t)\big)\bigg],
\label{eq:entropy_objective}
\end{equation}
where $\mathcal{H}(X) = \mathbb{E}[-\log p(x)]$ is Shannon's entropy, thereby punishing unnecessarily deterministic policies.  
The coefficient $\alpha$ balances the reward and the entropy terms, thereby giving some control over the stochasticity in the learned policy, however, this coefficient is usually annealed towards zero as training progresses.
The primary algorithm for MaxEnt RL is soft actor-critic (SAC)~\cite{haarnoja2018soft, haarnoja2018sac_aa}, which learns an on-policy, soft Q-function $Q^\pi_\text{soft}$ for an univariate Gaussian actor model. 
As can be seen in Fig.~\ref{fig:rollouts}, SAC learns one behavior (with local variations) but disregards other behaviors that reach the goal. 
This has two reasons. 
Firstly, SAC's actor model is unimodal and thus can not capture all possible modes, e.g. at the forks or intersections in the environment. While some prior works~\cite{tang2018implicit, haarnoja2017reinforcement} can learn multi-modal, entropy-regularized policies, multi-modality is not the key requirement to learning diverse behaviors. 
The second and more important reason why SAC disregards the other possible behaviors is that they are clearly sub-optimal, since their trajectories are longer and have higher costs compared to driving straight down the middle, from start to goal. 
RL is fundamentally only concerned with finding one optimal policy that solves the task, whether alternative solutions are possible is not considered, although this is crucial for transfer RL.
In the next section, we review methods that, unlike classical and MaxEnt RL, account for this and explicitly aim to learn diverse behavior alongside the optimal policy. 
\subsection{Learning diverse behaviors}
Popular approaches to learning diverse behaviors originate from unsupervised \textit{option}~\cite{sutton1999options} discovery~\cite{bacon2017option, fox2017multi, }. One such method is DIAYN~\cite{diyan2019}, which discovers distinct behaviors in an unsupervised manner and in the absence of a reward function, by maximizing the mutual information between behaviors and states~\cite{diyan2019}. 
Similarly, VALOR~\cite{valor2018} discovers distinct behaviors by maximizing the mutual information between behaviors and context vectors~\cite{valor2018}, again without access to a reward function. 
Both of these methods subsequently use the learned behaviors as low-level options in a hierarchical RL agent~\cite{sutton1999options} to solve downstream tasks efficiently. 

In this paper, we instead assume access to the reward function from the beginning and wish to exploit this information during learning, 
to discover alternative solutions to the given task.  
In this setting, \citeauthor{zhang2019novelPolicies}~\cite{zhang2019novelPolicies} learn multiple distinct policies for a task reward function $r_\text{task}$ by
training an autoencoder $\mathbf{D} = \{ \mathcal{D}_1, \dots, \mathcal{D}_n \}$ for each available policy $\pi_1, \dots, \pi_n$ on state sequences $\mathbf{S}_i = (\mathbf{s}_t, \mathbf{s}_{t+1}, \mathbf{s}_{t+K})$ of that policy and constructing a \textit{novelty} reward function
\begin{equation}
	r_\text{novel} = -\exp \big( -w \underset{\mathcal{D} \in \mathbf{D}}{\min} || \mathcal{D}(\mathbf{S}) - \mathbf{S} ||^2 \big).
\end{equation}
\citeauthor{zhang2019novelPolicies}~\cite{zhang2019novelPolicies} then update the policy using the angular bisector of the gradients on the expected novelty and task reward, to ensure that both objectives are improved.
Similarly, \citeauthor{zhou2022rewardSwitching}~\cite{zhou2022rewardSwitching} learn distinct policies for a given task by constraining policy search to trajectories $\tau$ that have low log-likelihood under already learned policies.
To promote diverse exploration, \citeauthor{zhou2022rewardSwitching}~\cite{zhou2022rewardSwitching} define, in addition to the extrinsic task reward $r^\text{ext}$, an intrinsic reward function $r^\text{int}$ based on learned, policy-specific reward models, to boost diverse exploration:
\begin{equation}
\bar{J}(\theta) = \mathbb{E}_{\tau \sim \pi_\theta}
	\bigg[ 
		\phi(\tau) \sum_{t=1}^{\infty} \gamma^{t-1} r_t^\text{ext} + \lambda \big(1 - \phi(\tau)\big) \sum_j r_t^\text{int}
	\bigg],
\label{eq:reward_switching_objective}
\end{equation}
where
\begin{equation}
	\phi(\tau) = \prod_{j=1}^{k-1} \mathbb{I} [\text{NLL}(\tau, \pi_j) \ge \delta]
\end{equation}
is an indicator function on negative log-likelihood of trajectories with threshold $\delta$.
While \cite{zhang2019novelPolicies, zhou2022rewardSwitching} exploit the task reward signal for learning novel policies for the given task, these methods either require learning additional novelty detectors, have to balance multiple reward signals or rely on expensive Monte Carlo updates.
In the next section, we propose a simple method 
for discovering alternative solutions for a given task,
while avoiding these shortcomings. 

\section{Learning contingent policies via novelty constraints}\label{sec:method}
\begin{figure}[t]
	\includegraphics[width=0.5\textwidth]{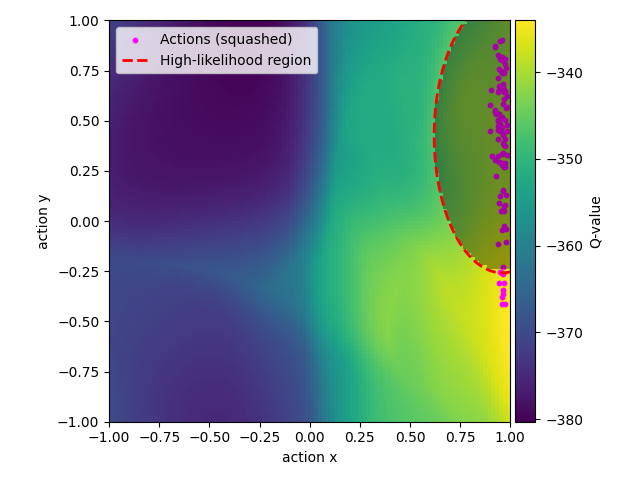}
	\centering
	\caption{Q-function, action samples, and high-likelihood region of the agent evaluated at the starting position. Novelty-constraints on policy-likelihood prevent the agent from using actions that fall into the high-likelihood region of prior policies.}
	\label{fig:indifference-space}
 \end{figure}
We propose an iteratively-constrained algorithm for learning alternative policies that can recover from contingent events. 
In each iteration, our algorithm learns a novel policy that attempts to solve the task, while its solution space is constrained to behavior that is unlikely under all previous policies for that task.  
Unlike ~\cite{zhang2019novelPolicies, zhou2022rewardSwitching}, we refrain from changing the agent's objective by introducing auxiliary novelty rewards, our agent still maximizes the expected task reward (subject to entropy-regularization), as in Eq.~\eqref{eq:entropy_objective}. This avoids the trade-off between the (potentially conflicting) objectives of maximizing return and behaving novel.
Instead and intuitively, to learn novel behavior, the agent should, in every state, only execute actions that are unlikely under prior policies for the same task. 
Following this intuition, we constraint policy search in the $i$-th iteration (i.e. learning of the $i$-th policy) to a set $\Pi_{i - 1}^\pi$ of policies, where policies in this set only select actions that are unlikely under all prior policies: 
\begin{equation}\label{eq:novelty-constrained-optimal-policy}
	\pi_i^* = \underset{\pi'}{\max}\  J(\pi') \mid \pi' \in \Pi_{i - 1}^\pi.
\end{equation}
To perform policy search as in Eq.~\ref{eq:novelty-constrained-optimal-policy}, the agent needs a way to sample actions from policies in $\Pi_{i - 1}^\pi$. Implementing Eq.~\eqref{eq:novelty-constrained-optimal-policy} locally and state-based, action selection for policies in $\Pi_{i - 1}^\pi$ is constrained:
\begin{equation}\label{eq:action_selection_constraint}
	\begin{split}
			& \ \ \ \ \ \ \ \ \ \ \ \ \ \ \ \ \ \ \ \ \mathbf{a} \sim \pi_i(\mathbf{s}) \\
			& \text{subject to}\ \ \ \pi_j(\mathbf{s}, \mathbf{a}) \le \varepsilon_j, \forall j \{ 1, \dots, i-1 \},
		\end{split}
\end{equation}
where $\varepsilon_j$ are thresholds specifying the maximally allowed action likelihood under policies from previous iterations $1, \dots, i-1$. Fig.~\ref{fig:indifference-space} provides a visualization of this  constraint and how it forbids actions that fall into the high-likelihood region of prior policies.
For each previous policy, we define an indicator function 
\begin{equation}
	\mathbb{I}_j^\pi(\mathbf{s}, \mathbf{a}) = 
	\begin{cases}
		1  & \text{ if } \pi_j(\mathbf{s}, \mathbf{a}), \le \varepsilon_j \\
		0 & \text{ otherwise},
	\end{cases}
\end{equation}
that can be used to check whether an action $\mathbf{a}$ in state $\mathbf{s}$ satisfies the \textit{novelty constraint} in Eq.~\eqref{eq:action_selection_constraint}. 
With the novelty constraint indicator functions we can project any policy into $\Pi_{i - 1}^\pi$:
\begin{equation}\label{eq:novelty-projection}
	\hat{\pi}_i(\mathbf{a} \mid \mathbf{s}) \propto \pi_i(\mathbf{a} \mid \mathbf{s}) \prod_{j=1}^{i-1} 	\mathbb{I}_j^\pi(\mathbf{s}, \mathbf{a}).
\end{equation}
Projecting policies into $\Pi_{i - 1}^\pi$ and sampling from $\hat{\pi}_i$ via rejection sampling is thus straightforward, however, we still require an algorithm for learning policies $\pi_i$ whose projections $\hat{\pi}_i$ perform well. We propose such a learning algorithm in the next section.

\subsection{Iterative novelty-constrained SAC}
In the $i$-th iteration of the novelty-constrained setting, the agent's true (novelty-constrained) policy is $\hat{\pi}_i$, to which we only have access via rejection sampling.
Thus, a learning algorithm for $\pi_i$ does not learn the agent's true policy but a proposal distribution for $\hat{\pi}_i$. 
To account for this, we propose an iterative and novelty-constrained version of SAC~\cite{haarnoja2018soft}.
Learning a critic for $\hat{\pi}_i$, the novelty-constrained policy, is straightforward by ensuring that the expectation of future actions in the TD-backup matches the (novelty-constrained) actor:
\begin{equation}\label{eq:novel_sac_critic_update}
	J_Q(\theta_i) = \mathbb{E}_{\mathbf{s}_t, \mathbf{a}_t, \mathbf{s}_{t+1} \sim \mathcal{D}} \bigg[
	\frac{1}{2} 
	\big(
	Q_{\theta_i}(\mathbf{s}_t, \mathbf{a}_t) - 
	\hat{Q}_{\theta_i}(\mathbf{s}_t, \mathbf{a}_t)
	\big)
	\bigg],
\end{equation}
with
\begin{equation}
	\begin{aligned}
		\hat{Q}_{\theta_i}(\mathbf{s}_t, \mathbf{a}_t)& = 
		r(\mathbf{s}_t, \mathbf{a}_t) 
		+ 
		\gamma \\
		& 
		\mathbb{E}_{\mathbf{a}_{t+1} \sim \hat{\pi}_i}
		\big[
		Q_{\bar{\theta_i}}(\mathbf{s}_{t+1}, \mathbf{a}_{t+1})
		-
		\log(\pi_i(\mathbf{a}_{t+1} \mid \mathbf{s}_{t+1}))
		\big],
		\label{eq:test}
	\end{aligned}
\end{equation}
where $\bar{\theta_i}$ refers to the target network parameter for $\hat{\pi}_i$'s critic. 
Similarly, the actor update must reflect the novelty constraint and rejection sampling step as well. 
A key property of SAC is that it updates the actor by minimizing the KL divergence between the actor and the critic: 
\begin{equation}
	\begin{split}
	J_\pi(\phi) & = 
	\mathbb{E}_{\mathbf{s}_t \sim \mathcal{D}} 
	\bigg[
	\text{D}_{\text{KL}}
	\infdiv*
	{\pi_\phi(\cdot \mid \mathbf{s}_t)}
	{\frac{\exp(Q_{\theta_i}(\mathbf{s}, \cdot))}{Z_{\theta_i}(\mathbf{s}_t)}}
	\bigg] \\
	& = \mathbb{E}_{\mathbf{s}_t \sim \mathcal{D}, \mathbf{a}_t \sim \pi_\phi} 
	\big[
		\log \pi_\phi(\mathbf{a}_t \mid \mathbf{s}_t) 
		- 
		Q_{\theta_i}(\mathbf{s}_t, \mathbf{a}_t)
	\big].
	\end{split}
\end{equation}
In our case, we can still backpropagate through the critic, however, only for action samples that have low likelihood under previous policies and satisfy the novelty constraint (i.e. for which $\prod_{j=1}^{i-1} \mathbb{I}_j^\pi(\mathbf{s}, \mathbf{a}) = 1$). 
Action samples that violate the novelty constraint, i.e. are likely under previous policies, should follow a different gradient because they would be discarded by the rejection and sampling step. 
Since these actions are never executed, the critic never observes a reward signal for those actions and hallucinates unreliable value estimates whose gradients are not suited for learning the novelty-constrained actor. 
Thus, to account for the rejection sampling step and to encourage learning an actor that respects the novelty constraint in Eq.~\eqref{eq:action_selection_constraint}, for actions that violate any of the $i-1$ constraints, 
we instead use the gradient of the KL divergence between the current policy $\pi_i$ and the policies whose constraints are violated. This leads to the following actor update for the proposed, novelty-constrained SAC:
\begin{equation}\label{eq:novel_sac_actor_update}
	\begin{split}
		J_\pi(\phi_i) & = \mathbb{E}_{\mathbf{s}_t \sim \mathcal{D}, \mathbf{a}_t \sim \pi_{\phi_i}} \\
		& \bigg[ 
		\prod_{j=1}^{i-1} \mathbb{I}_j^\pi(\mathbf{s}, \mathbf{a}) 
		\overbrace{
			\log \pi_{\phi_i}(\mathbf{a}_t \mid \mathbf{s}_t) 
			- 
			Q_{\theta_i}(\mathbf{s}_t, \mathbf{a}_t)
		}^{\text{D}_\text{KL}(\pi_i \mid\mid Q)} + \\
		& 
		\bigg( 
		1 - \prod_{j=1}^{i-1} \mathbb{I}_j^\pi(\mathbf{s}, \mathbf{a}) 
		\bigg)
		\underbrace{
			\log \pi_{\phi_i}(\mathbf{a}_t \mid \mathbf{s}_t) 
			- 
			\log \pi_{\phi_j}(\mathbf{a}_t \mid \mathbf{s}_t)
		}_{\text{D}_\text{KL}(\pi_i \mid\mid \pi_j)} \bigg].
	\end{split}
\end{equation}
Thus, our iterative algorithm for learning contingent behaviors operates as follows. The first policy $\pi_1$ is learned unconstrained, using normal SAC~\cite{haarnoja2018composable}, thus $\pi_1$ is the optimal soft policy for the given task. 
Once $\pi_1$ has converged, our algorithm proceeds with the learning of $\pi_2$, which is novelty-constrained w.r.t $\pi_1$, meaning it has to solve the given task as best as possible while respecting the constraint in Eq.~\eqref{eq:action_selection_constraint}. $\pi_2$ is learned using the critic update in Eq.~\eqref{eq:novel_sac_critic_update}, the actor update in Eq.~\eqref{eq:novel_sac_actor_update} and relies on rejection sampling to generate actions from $\hat{\pi}_2$. Once $\pi_2$ has converged, $\pi_3$ can be learned, being novelty constrained w.r.t $\pi_1$ and $\pi_2$, and so on. 
In the next section, we show how these additional policies can be used to recover from unforeseen events in the transfer task.

\section{Recovering from unforeseen events using contingency policies}

\begin{figure}
	\includegraphics[width=0.5\textwidth]{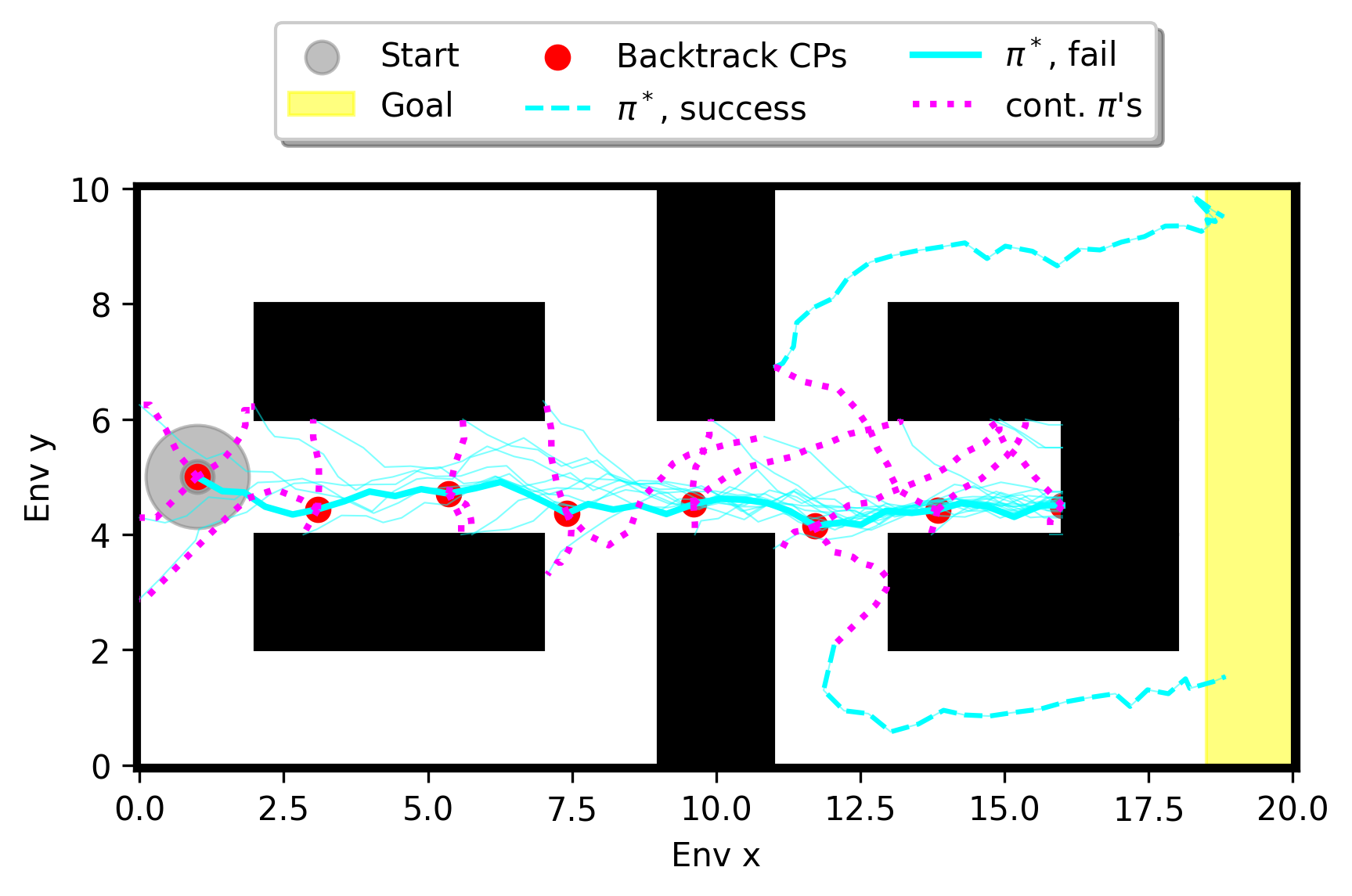}
	\centering
	\caption{Rollouts generated by our backtracking algorithm. The algorithm
attempts to recover by rolling out the available contingency policies, followed by rollouts of the optimal policy. If unsuccessful, the agent backtracks to the next checkpoints and executes the contingency policies and optimal policy again. This repeats until the task is finished successfully or all checkpoints are exhausted.}
	\label{fig:contingency-results}
\end{figure}
When we transfer a pre-trained agent to a new task, the agent can be exposed to situations that require it to deviate from its behavior learned during pre-training. For example, one such event might be when the middle path in Fig.~\ref{fig:rollouts} becomes blocked. 
Our proposed method accounts for this by learning additional policies during pre-training, to be used in and recover from such situations. To know when we should use one of the contingency policies instead of the optimal policy, we require a method for detecting contingent events, where the optimal pre-trained policy behaves sub-optimally. In the scope of this workshop paper, we simply rely on $\Delta \mathbf{s}$, i.e.changes in the state variable, to detect such events and leave a more sophisticated method as future work. 
\begin{figure}
	\includegraphics[width=0.5\textwidth]{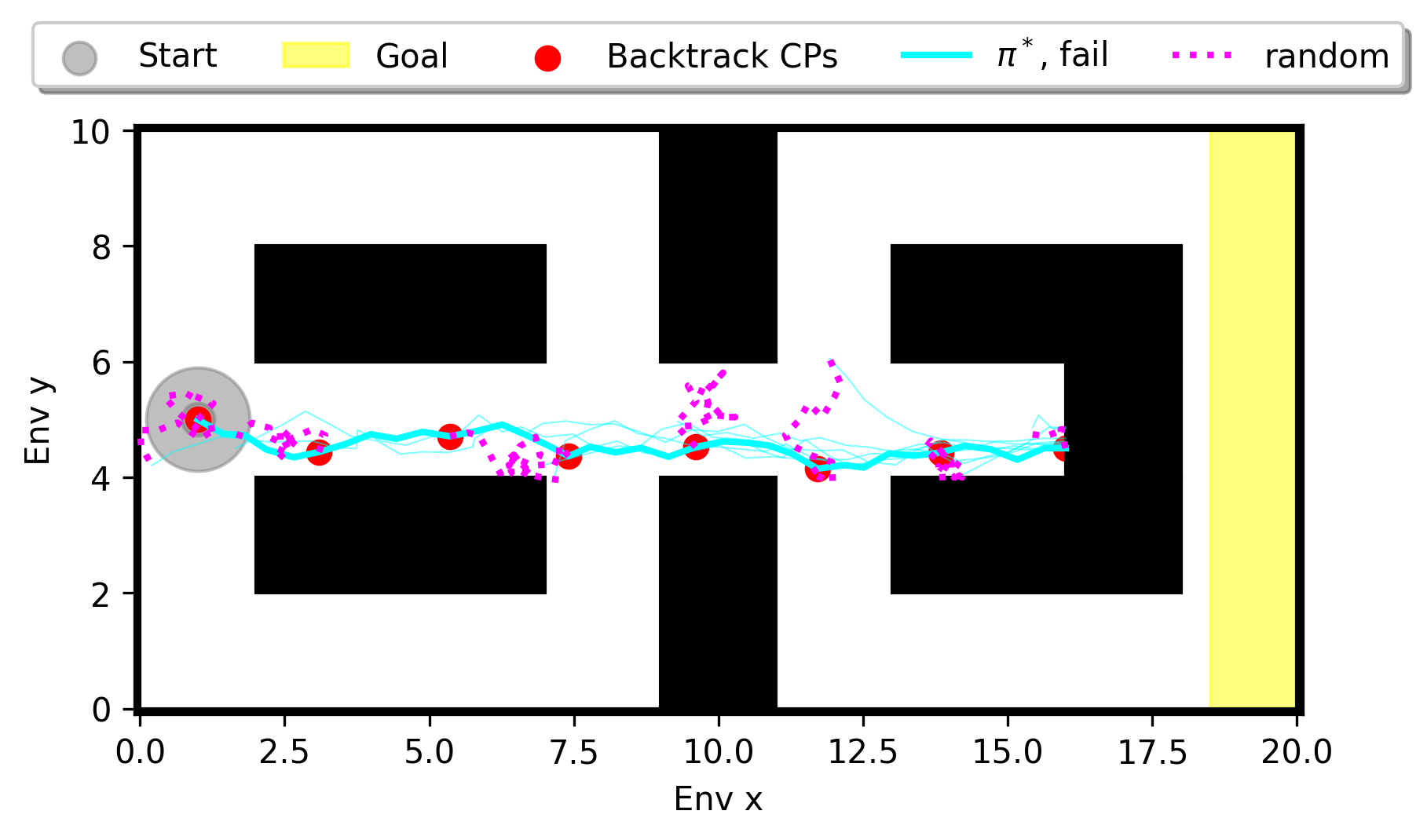}
	\centering
	\caption{Rollouts generated by our backtracking algorithm but with random behavior instead of contingent recovery policies.}
	\label{fig:random-results}
\end{figure}
%
Given we detect a contingency, the agent iterates between backtracking for $k$ steps, running a contingency policy for $m$ steps and then rolling out the optimal policy. This uninformed process does not require additional knowledge, e.g. a model of environment transition dynamics, and works well in practice, as seen in Figure~\ref{fig:contingency-results}. This is in contrast to a baseline comparison in Fig.~\ref{fig:random-results}, where the agent executes random actions to recover from the contingencies, instead of the recovery policies learned with our proposed method.

There are a number of points that should be addressed in future work. Instead of treating $k$ and $m$ as hyperparameters, it would be preferable to automatically identify states to backtrack to and automatically decide for how long and which contingency policy to execute. For the result in Fig.~\ref{fig:contingency-results}, we manually selected values that were adequate for our simple testing environment, which is not practical for more sophisticated problems.
We leave these points, as well as more thorough experimentation and baseline comparisons, as important future work.
\addtolength{\textheight}{-0cm}   

\printbibliography

\end{document}